\documentclass[11pt]{article}

\usepackage[final]{acl}

\usepackage{times}
\usepackage{latexsym}

\usepackage[T1]{fontenc}

\usepackage[utf8]{inputenc}

\usepackage{microtype}

\usepackage{inconsolata}

\usepackage{graphicx}

\usepackage{multirow}
\usepackage{amsmath}
\usepackage{tabularx}
\usepackage{caption}
\usepackage{algorithm2e}
\RestyleAlgo{ruled}
\usepackage{amssymb}
\usepackage{subcaption}
\usepackage{xcolor}
\usepackage{hypcap}
\usepackage{needspace}
\usepackage{booktabs}
\usepackage[stable]{footmisc}
\usepackage{hyperref}

\newcommand{\mathbbm}[1]{\text{\usefont{U}{bbm}{m}{n}#1}}
\newcommand{\model}{\leavevmode{\fontfamily{qag}\selectfont\textbf{\textcolor{teal!80!black}{\textsc{MagiC}}}}}
\newcommand{\score}{{\fontfamily{qag}\selectfont \textsc{MagiScore}}}

\title{\model: Evaluating Multimodal Cognition \\ Toward Grounded Visual Reasoning}

\author{%
  Chengfei Wu\thanks{Equal contribution.} \\
  Purdue University \\
  \texttt{wu1491@purdue.edu} \\
  \And
  Ronald Seoh$^*$ \\
  Purdue University \\
  \texttt{bseoh@purdue.edu} \\
  \And
  Bingxuan Li \\
  Univ. of Illinois Urbana-Champaign \\
  \texttt{bl61@illinois.edu} \\
  \AND
  Liqiang Zhang \\
  University of Connecticut \\
  \texttt{liqiang.zhang@uconn.edu} \\
  \And
  Fengrong Han \\
  University of California Berkeley \\
  \texttt{bohanmia@berkeley.edu} \\
  \And
  Dan Goldwasser \\
  Purdue University \\
  \texttt{dgoldwas@purdue.edu} \\
}

\begin{document}

\maketitle

\begin{abstract}
Recent advances in large vision-language models have led to impressive performance in visual question answering and multimodal reasoning. However, it remains unclear whether these models genuinely perform grounded visual reasoning or rely on superficial patterns and dataset biases. In this work, we introduce \model, a comprehensive benchmark designed to evaluate grounded visual reasoning over natural scenes---assessing not only answer accuracy but also the quality of step-by-step reasoning and its alignment with relevant visual evidence. Our benchmark includes 5,534 weakly supervised QA examples generated from strong model outputs and 896 human-curated examples with fine-grained annotations, including answers, rationales, and bounding box groundings. We evaluate 15 vision-language models ranging from 7B to 70B+ parameters across four dimensions: final answer correctness, reasoning validity, grounding fidelity, and self-correction ability. \model{} further includes diagnostic settings to probe model robustness under adversarial visual cues and assess their capacity for introspective error correction. We introduce new metrics such as \score{} and StepSense, and provide comprehensive analyses that reveal key limitations and opportunities in current approaches to grounded visual reasoning over natural scenes.
\end{abstract}

\section{Introduction}
\label{sec:01-introduction}

Recent advances in large vision-language models (LVLMs) have substantially improved visual question answering and reasoning. Proprietary models such as GPT-4V~\citep{GPT4V}, Gemini~\citep{Gemini}, and Claude-3~\citep{Claude}, along with open-source counterparts like LLaVA~\citep{NEURIPS2023_6dcf277e}, Qwen-VL~\citep{Qwen-VL}, and DeepSeek-VL~\citep{lu2024deepseekvl}, exhibit strong capabilities in perception, object recognition, and complex visual understanding. However, these successes raise a deeper question: \textit{Do current multimodal models genuinely reason over visual content, or do they rely on superficial patterns and dataset biases?}

\begin{figure*}[htpb]
\centering
\includegraphics[width=1.95\columnwidth]{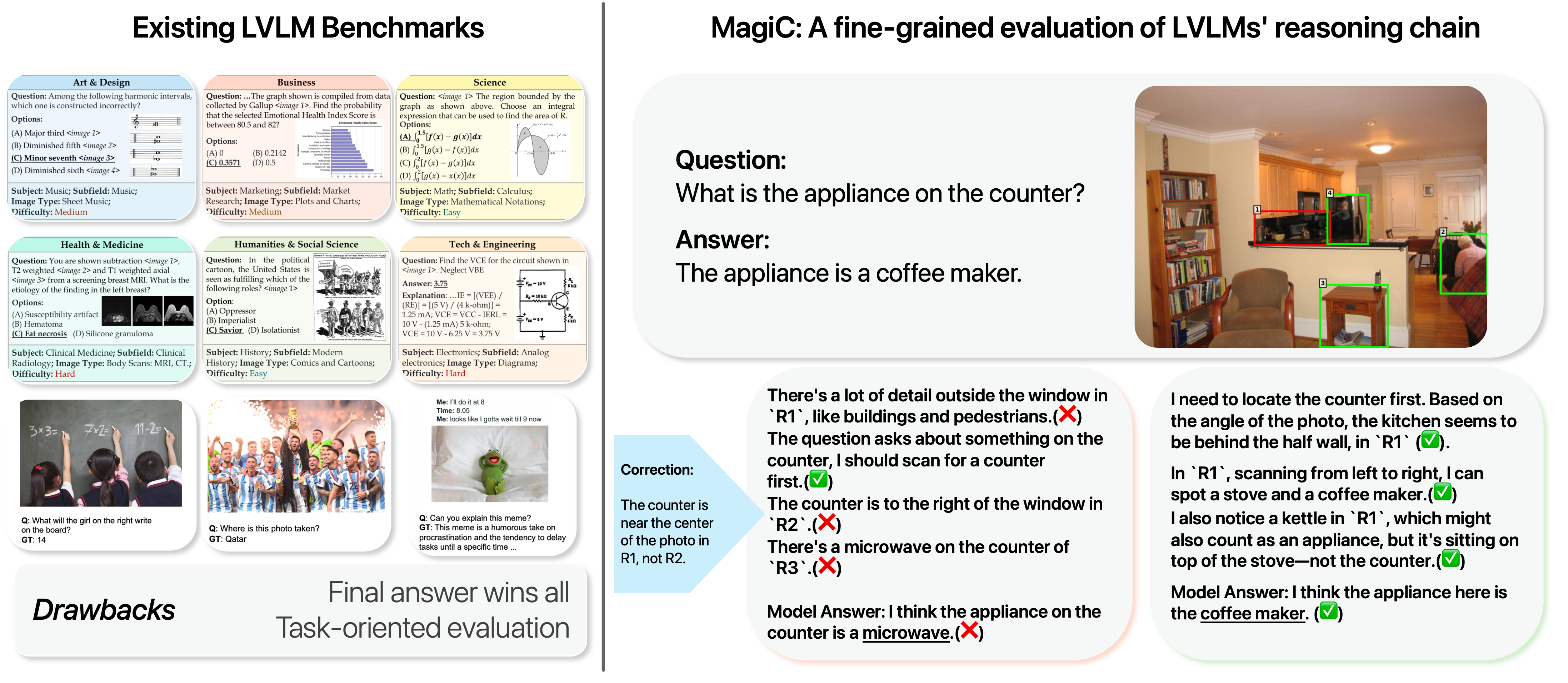}
\vspace*{-0.8em}
\caption{Example scenarios from \model{}. Existing work mainly focuses on evaluating the final answer for a given task ignoring the steps model takes to answer.}
\vspace*{-1em}
\label{fig:reasoning-example}
\end{figure*}

Addressing this question requires a closer examination of grounded visual reasoning over natural scenes---the ability to selectively attend to relevant visual inputs and integrate them into coherent, multi-step logical reasoning. This ability is central to human intelligence and is critical for building interpretable, trustworthy AI systems. For instance, as illustrated in our example in \autoref{fig:reasoning-example}, robust reasoning requires the model to justify its answer by explicitly referencing the correct visual regions and following a logical inference path. 

Yet, most existing benchmarks primarily focus on end-task performance, offering limited insight into whether models truly "understand" the image or simply exploit dataset biases. They rarely examine how models arrive at answers, whether intermediate reasoning steps align with relevant visual regions \citep{li2023seedbenchbenchmarkingmultimodalllms,10.1007/978-3-031-72658-3_13,Yue_2024_CVPR,10.5555/3692070.3694451} Recently, increasing attention has been given to the reasoning processes of VLMs, with several works examining their chain-of-thought reasoning~\citep{zhang2024improve, zhao2025cotvla, chen2023see}, intermediate rationale generation~\citep{chen2023see}, and visual grounding consistency~\citep{yang2023improving, he2024improved}. However, these efforts remain fragmented, often lacking unified evaluation protocols and comprehensive grounding diagnostics.

In this work, we introduce a comprehensive benchmark, \model{} (Bench\textbf{MA}rkin\textbf{G} Mult\textbf{I}modal \textbf{C}ognition), designed to evaluate grounded visual reasoning over natural scenes. As illustrated in the right section of \autoref{fig:reasoning-example}, our framework goes beyond conventional question answering by measuring a model’s ability to (1) generate correct final answers, (2) articulate coherent, interpretable step-by-step reasoning, (3) ground each reasoning step in the appropriate visual evidence, and (4) intervene upon and revise faulty reasoning when possible.


To support this evaluation, we construct a dual-sourced dataset consisting of 5,534 weakly-supervised visual reasoning instances and 896 human-curated instances with over 15,000 annotated reasoning steps. The weakly supervised data is automatically derived from the outputs of high-performing vision-language models across $78$ diverse question types, enabling broad coverage with minimal annotation cost. In contrast, the human-curated set includes fine-grained annotations—answers, reasoning rationales, and bounding box groundings—with a held-out test set of $698$ instances for rigorous evaluation.

We evaluate 15 vision-language models ranging from 7B to 70B+ parameters, including both open-source and proprietary systems. Our evaluation framework captures four dimensions of model performance: short-form and long-form answer correctness, reasoning validity, grounding quality, and self-correction ability. Grounding quality is quantified using \score{}, which measures the overlap between predicted and reference bounding boxes. To assess models’ introspective capabilities, we further analyze their ability to detect and revise intermediate reasoning errors.

In addition to standard performance metrics, \model{} introduces several diagnostic settings that probe robustness and interpretability. The adversarial grounding setting evaluates models under misleading or irrelevant visual cues to test their reliance on correct evidence. The self-correction analysis examines whether a model can recognize and revise its own reasoning errors. Finally, we conduct a human evaluation of reasoning quality on a subset of four models, deriving the \textit{StepSense} score that captures coherence and factual consistency.


In summary, our contributions are as follows: 

\begin{itemize}
    \item First, we introduce a novel task that jointly evaluates answer correctness, step-by-step reasoning quality, and visual grounding fidelity.
    
    \item Second, we construct a benchmark dataset comprising 5,534 weakly supervised interleaved bounding-box QA examples and 896 high-quality human-curated examples with detailed annotations for answers, rationales, and bounding boxes.

    \item Third, we introduce novel evaluation metrics, \score{} tailored to grounded visual reasoning over natural scenes, with \textit{StepSense} and \textit{Self-Heal} which measures model reasoning quality and self-correction ability.
    
    \item Lastly, we conduct a comprehensive analysis of 15 state-of-the-art models, uncovering key insights into both the limitations and potential of current approaches to grounded visual reasoning. Our findings indicate that models exhibiting precise region focus are generally more likely to answer questions correctly.
\end{itemize}

We hope this benchmark will catalyze future research toward building more interpretable, robust, and cognitively aligned multimodal systems.

\section{Related Work}
\label{sec:02-related_works}
\subsection{Benchmarks for Large Vision-Language Models}

Ever since the introduction of large vision-language models (LVLMs) such as BLIP \citep{li2023blip2bootstrappinglanguageimagepretraining} and LLaVA \citep{NEURIPS2023_6dcf277e}, the field has seen rapid advancements in both model scaling and evaluation methodologies. Researchers have increasingly focused on scaling models at test time to push the boundaries of what these systems can achieve. Alongside these developments, there has been a notable surge in the creation of new benchmarks specifically designed to assess the visual reasoning capabilities of LVLMs, moving beyond traditional vision tasks like Visual Question Answering (VQA). Recent benchmarks such as MMBench \citep{10.1007/978-3-031-72658-3_13}, MMMU \citep{Yue_2024_CVPR}, SEED-Bench \citep{li2023seedbenchbenchmarkingmultimodalllms}, and MM-Vet \citep{10.5555/3692070.3694451} now provide more comprehensive and fine-grained evaluations, covering a wider variety of visual reasoning challenges. However, most of these benchmarks still primarily measure end task performances and do not delve deeply enough into the quality of LVLMs' reasoning process. Despite efforts on visual chain of thought (CoT) \citep{chen2023see,rose2024visualchainthoughtbridging, zhang2024improve, zhang2024multimodal, chen-etal-2024-measuring,menon2024whiteboardofthoughtthinkingstepbystepmodalities,shao2024visualcotadvancingmultimodal,zhao2025cotvla,wu2025viscobenchmarkingfinegrainedcritique,xu2025llavacotletvisionlanguage,thawakar2025llamavo1rethinkingstepbystepvisual}, evaluating their explicit reasoning quality remains an extremely challenging problem and the current methods leave substantial room for improvement. 

\subsection{Grounding in Vision Tasks}

In order to address the gap in existing benchmarks, we examine in particular how well textual reasoning steps remain \emph{grounded in visual input}. Visual grounding has typically been studied in relation to the internal activations of vision encoders and/or in the context of specific visual task \citep{10.1007/978-3-031-19833-5_38, yang2023improving, he2024improved, reich-schultz-2024-uncovering, NEURIPS2024_f38cb4cf, 10.1007/978-3-031-72986-7_12,NEURIPS2023_6dcf277e} However, our work examines visual grounding more generally in the context of LVLMs' explicit reasoning - if the textual reasoning loses its connection to visual reality, the generated explanations and answers, no matter how fluent, may be incorrect or irrelevant. Our work aims to provide methods for tracing and evaluating this connection throughout the reasoning process.

\subsection{Grounded Visual Reasoning Methods}

Recent training methods for improving grounding, such as Visual-RFT \citep{liu2025visualrft}, ViGoR \citep{yan2024vigor}, UniVG-R1 \citep{bai2025univgr1}, and VGR \citep{wang2026vgr}, are the class of approaches \model{} is designed to measure: our region-reference fine-tuning experiment demonstrates precisely this usage, lifting \score{} macro-F1 from 43.2 to 82.2.

\section{The \model{} Benchmark}
\label{sec:03-dataset}
We introduce \model{}, the first benchmark dataset specifically designed to evaluate a model's ability to selectively attend to relevant visual information and integrate it into coherent, multi-step logical reasoning. \model{} contains $\mathbf{698}$ test question-and-reasoning pairs, each densely annotated with step-wise correctness labels and corresponding natural-language corrections. The test set encompasses $\mathbf{41}$ distinct question subtypes derived from the GQA dataset \citep{hudson2018gqa}, categorized into five primary question types: 1) \textbf{Verify}: Binary yes/no questions. 2) \textbf{Query}: Open-ended questions. 3) \textbf{Choose}: Questions offering a choice between two alternatives (e.g., “Is it red or blue?”). 4) \textbf{Logical}: Questions requiring logical inference. 5) \textbf{Compare}: Questions involving comparisons between two or more objects. 

\begin{figure}[!htbp]
    \centering
    \includegraphics[width=150pt]{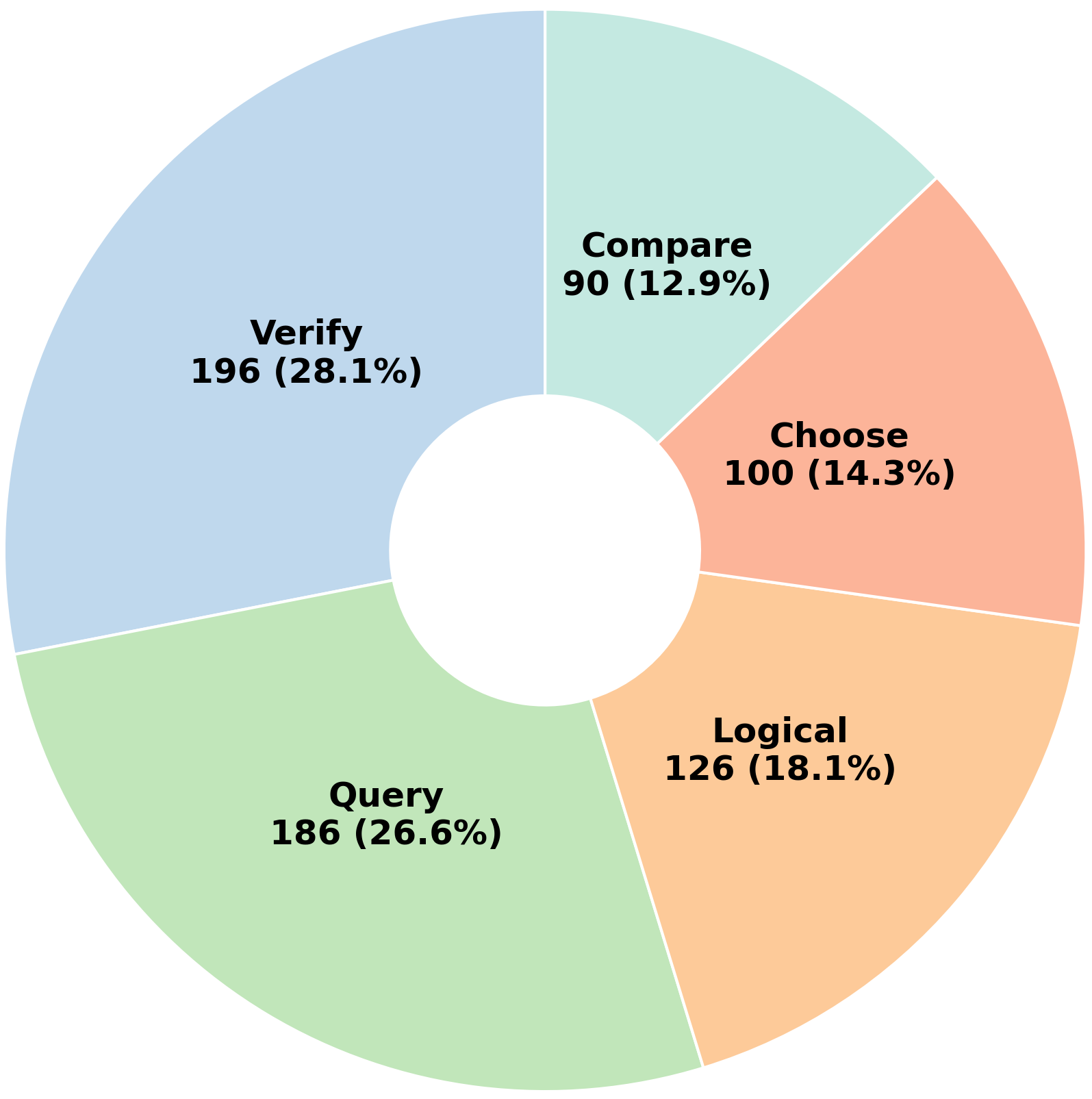}
    \captionof{figure}{Question type distributions. \textit{Detailed distribution of detailed types can be found in Figure \ref{fig:type-distribution-full}.}}
    \label{fig:type-distribution-small}
    \vspace{-1em}
\end{figure}

Detailed statistics of the dataset are summarized in \autoref{tab:data-stat}. In total, we collected $896$ annotated questions, comprising correctness annotations for $8403$ individual reasoning steps and $2700$ natural-language corrections. In which, $198$ are reserved for development and $698$ for test, and the test set is used as the \model{} benchmark. The distribution of the question types of the test set can be found at Figure \ref{fig:type-distribution-small}.

The construction of \model{} involves three main steps: task-input processing, response collection, and human correction collection. Examples of the dataset, are available in the \autoref{sec:app-data-example}.

\subsection{Task-Input Construction} \label{sec:03-dataset-task-input}
We obtain images ($I$), questions ($Q$), and short/full ground-truth answers ($GT_{short}$, $GT_{full}$) from the GQA dataset \citep{hudson2018gqa}. We restrict our selection to the public \texttt{val} splits aim to minimize the likelihood of dataset overlap with existing model knowledge. Additionally, we incorporate human saliency maps ($Saliency_{q}$) from AiR-D \citep{air2020}, which are based on human eye-tracking data, providing higher quality saliency regions compared to pointer-based counterparts. Although AiR-D contains saliency maps corresponding to both correct and incorrect human answers, we only utilize those maps associated with correct responses.

\begin{figure}
    \centering
    \vspace{-1em}
    \includegraphics[width=1.0\linewidth]{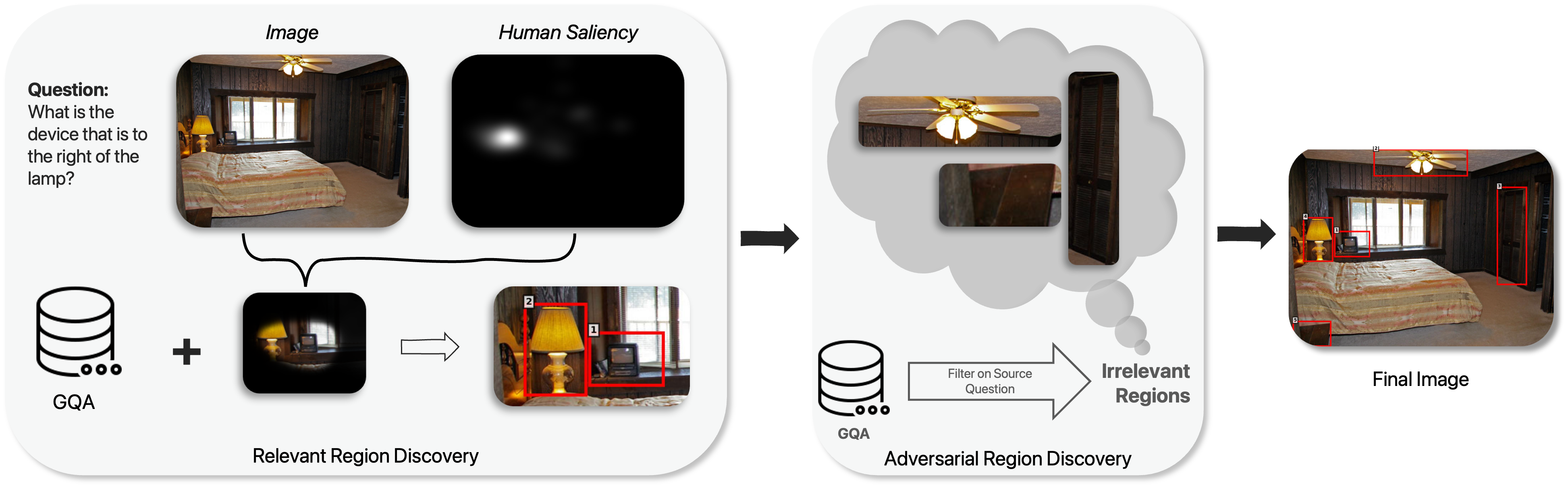}
    \caption{Illustration of Task-Input Data Construction Processing. \textit{Human saliency maps are used only during construction to identify relevant regions and are never provided to LVLMs at evaluation time.}}
    \vspace*{-1.5em}
    \label{fig:saliency-demo}
\end{figure}

As illustrated in the relevant region discovery section in \autoref{fig:saliency-demo}, for each question $Q$ and its associated saliency map $Saliency_{q}$, we identify candidate “hotspot” regions that received significant human attention. Bounding boxes for these regions are extracted and merged with the annotated objects from GQA, forming an initial set of relevant regions. These regions are further processed to eliminate overlaps or merge boxes where necessary, resulting in the final set of relevant bounding boxes, $RBox_{q}$, for question $Q$.

Next, building upon the identified relevant regions ($RBox_{q}$) and GQA's image-level object annotations, we select additional image regions relevant to other GQA questions but irrelevant to the current question $Q$, as demonstrated in \autoref{fig:saliency-demo}. These are termed adversarial regions. We randomly sample adversarial regions that do not overlap with any region in $RBox_{q}$, creating the set $ABox_{q}$. To ensure sufficient adversarial content, we sample three adversarial regions per question; if fewer than three such regions exist, additional regions are generated randomly. We henceforth refer to $Box_{q} = RBox_q \cup ABox_{q}$ as the candidate box set and to the labels indicating which boxes are relevant as the ground-truth relevance. A detailed implementation can be found in \autoref{alg:pre-process}.

This step yields tuples of the form $(I_q, Box_q, Q_q, A_q)$ for each question $q$.

\subsection{LVLM Response Collection}
\label{sec:03-dataset-response-collection}

Using the processed input data, we collect responses from four LVLMs representing four different model families: InternVL-2.5 8B, Gemma 3 27B, Qwen2.5-VL 7B, and OpenAI GPT-4o. Each model is prompted using in-context learning (ICL) examples created in-house, provided with inputs $(I_q, Q_q)$, and tasked to generate step-by-step reasoning from the first-person perspective, concluding with a final answer.

Responses exhibiting invalid formatting or repetitive reasoning loops are filtered out. Remaining valid responses are segmented into individual sentences, each representing a distinct reasoning step $S_i$. From these valid responses, we randomly select one reasoning sequence per question ($S_{q}$), which is then densely annotated by human annotators.

\subsection{Reasoning Correction Collection}
\label{sec:03-dataset-annotation}
We recruit four expert annotators (computer science students with relevant experience) to evaluate and correct the sampled model responses. Annotators begin by annotating ten practice examples, which serve as training materials and are discarded thereafter. In the annotation phase, annotators first verify the quality of each question and compare the model's final answer against the ground truth. Subsequently, they assess the correctness of each individual reasoning step $S_i$. This annotation archives a substantial inter-annotator agreement of 0.72 Cohen's Kappa. Annotators are required to document any ambiguities, poorly formulated questions, or reasoning steps requiring reconsideration. A discussion session is conducted after every 100 annotated examples to resolve disagreements and collectively refine annotations.

For each incorrect reasoning step $S_i$, annotators will also provide a corrected natural-language version. These corrections are complete rewrites rather than simple critiques, intended to maintain coherent and logical reasoning when replacing the original step.

After completing this annotation step, each question $Q$ is associated with an annotated reasoning sequence $S_{q} = \{s_1, s_2, \ldots, s_n\}$, where each element $s_n$ consists of the original reasoning step, a correctness label, and its corresponding correction $s_n= (original, correctness, correction)$.

\subsection{Weakly Supervised Dataset}

In addition to the human-annotated dataset, we also created a weakly supervised dataset following the pipeline similar to that described in Section~\ref{sec:03-dataset-task-input}. The weakly-supervised split omits human saliency but derived its boxes from GQA’s object bounding boxes. To avoid overlap with our test set, images included in this dataset are distinct from those in the primary annotated set. Model responses in this version were collected using inputs $(I_q, Q_q, A_q)$ without adversarial regions to produce cleaner reasoning data. We also conducted a training run using this split, detailed results are reported in \autoref{sec:app-weak-supervision}.

\section{Task Overview}
\label{sec:04-task_formulation}
Using an image-question pair $(I, Q)$ from the \model{} benchmark, we sample model responses from Large Vision-Language Models (LVLMs), denoted by $S_{\mathcal{LM}}$. The response from a given model can be represented as $S_{\mathcal{LM}} = [s_1, s_2, \dots, s_n]$, where $n$ is the number of reasoning steps generated by the model, along with the model's final answer $A_{\mathcal{LM}}$. Additionally, the dataset provides ground-truth bounding boxes $Box_q$ and corresponding ground-truth answers in both short and long form, denoted by $(GT_{short}, GT_{full})$. Each question $Q$ also includes a human-corrected reasoning chain, $S_{q}$. Utilizing the tuple $(I, Q, Box_q, S_{\mathcal{LM}}, A_{\mathcal{LM}}, GT_{short}, GT_{full}, S_{q})$, we evaluate LVLMs on the following three tasks:

\subsection{Region Focus} \label{sec:formulation-attention}

For each question $Q$, a LVLM generates a reasoning chain $S_{\mathcal{LM}}=[s_{1}, s_{2},\dots , s_{n}]$. We introduce the region extractor function $\phi(\cdot)$ that maps each reasoning step $s_i$ to a set of bounding box indices from $Box_q$, defined as $\phi(s_{i})\subseteq\{1,\dots ,|Box_q|\}$. By taking the union of these regions across all reasoning steps, we derive the complete set of regions the model focused on as part of its reasoning: $\hat{\mathcal{R}}=\bigcup_{i=1}^{n}\phi(s_{i})$.

We can then represent the set of unique regions explored by the model as a binary vector:
$\hat{\mathbf{y}}\in\{0,1\}^{|Box_q|}, \text{where} \quad \hat{\mathbf{y}}_{k}=\mathbbm{1}[k\in\hat{\mathcal{R}}]$. Here, $\hat{\mathbf{y}}_{k}=1$ indicates that at least one reasoning step utilizing the $k$-th bounding box. Analogously, we have the ground-truth vector from the benchmark: 
$\mathbf{y}^{*}\in\{0,1\}^{|Box_q|}, \text{where} \quad \mathbf{y}^{*}_{k}=1$ if and only if the $k$-th box in $Box_q$ is labeled as relevant.

Using $\hat{\mathbf{y}}$ and $\mathbf{y}^{*}$, we evaluate two variants of region-focus scores:

\score{} (micro): Aggregating predictions across all images, we calculate precision, recall, and $F_1$ scores based on cumulative totals of true positives, false positives, and false negatives.

\score{} (macro): For each question $Q$, we first compute image-level precision, recall, and $F_1$ individually. The macro-level score is then obtained by averaging these metrics across all questions.

Note that the primary goal of \score{} is to assess visual grounding and reasoning capabilities independently of object detection performance, making the metrics distinct from the detection-oriented measures such as Intersection over Union (IoU). Detailed illustration of the score calculation is in \autoref{sec:app-score-example}.

\paragraph{Region extraction and metric scope.}
The region extractor \(\phi(\cdot)\) maps each textual mention of a numbered candidate box to its corresponding index. We emphasize that \score{} measures \emph{explicit} evidence selection against human-validated relevant and adversarial boxes. It is not a measure of a model's internal attention, nor of the faithfulness of each reasoning step. StepSense evaluates step-level faithfulness, and the two metrics are complementary components of the same protocol.

\subsection{LLM-assisted Final Answer Accuracy}

While \citet{hudson2018gqa} propose exact match as the evaluation metric for answer accuracy, this approach has notable limitations when evaluating natural language responses generated by LVLMs. Specifically, exact match fails to recognize synonymous or semantically equivalent responses (e.g., "tan" vs. "brown") and may incorrectly reward random occurrences of the ground-truth term within longer responses. Therefore, we propose an LLM-assisted evaluation approach by leveraging a large language model as a judge to more effectively assess correctness.

The LLM-based judge is modeled as the following: $f_{\mathcal{LM}}: P(A_{\mathcal{LM}},GT_{full}) \mapsto \{0,1\}$

This function classifies a model's predicted answer $A_{\mathcal{LM}}$ against a provided ground-truth answer. Using this function, we define two accuracy measures:

\textbf{1. Short Answer Accuracy} (\textbf{$\text{Acc}_{\text{short}}$}): evaluates whether the predicted answer captures essential elements indicated by the short ground-truth answer: $\text{Acc}_{\text{short}} = \frac{1}{|\mathcal{D}|}\sum_{(I,Q)\in\mathcal{D}} f_{\mathcal{LM}}\bigl(A_{\mathcal{LM}}, GT_{\text{short}}\bigr)$

\textbf{2. Full Answer Accuracy} (\textbf{$\text{Acc}_{\text{full}}$}): evaluates whether the predicted answer aligns comprehensively with the detailed ground-truth answer: $\text{Acc}_{\text{full}} = \frac{1}{|\mathcal{D}|}\sum_{(I,Q)\in\mathcal{D}} f_{\mathcal{LM}}\bigl(A_{\mathcal{LM}}, GT_{\text{full}}\bigr)$

The short-answer accuracy measure can be viewed as a more permissive evaluation metric compared to full-answer accuracy, as it only requires capturing core semantic elements identified by the concise ground-truth answer.

\paragraph{Evaluation mechanisms.}
\model{} uses three distinct evaluation mechanisms: (1) answer accuracy is judged by an LLM via reference-anchored semantic equivalence; (2) StepSense is based on fully human-annotated step-level labels; and (3) Self-Heal uses an LLM to decide whether a model continuation semantically matches a human-written correction. These mechanisms are validated separately in \autoref{sec:app-judge-validation}.

\subsection{Self-Correction} \label{sec:formulation-self-corr}
For each question $Q$, the benchmark provides human-corrected reasoning chains, denoted as $S_q = [s_{1}, s_{2}, \dots, s_{n}]$, where each step has been labeled with correctness and natural language correction if is incorrect. Given the annotated reasoning chain $S_{q}$, we identify contiguous subsequences of incorrect reasoning steps marked by human annotators as $S_{wrong} = [s_i, \dots, s_j]$, with indices $2 \leq i \leq j \leq n - 2$. 

To evaluate the self-correction ability of the LVLM, we design an intervention experiment. Specifically, we inject the initial reasoning steps up to and including the incorrect step, i.e., the subsequence $[s_1, s_2, \dots, s_j]$, into model's response. With this partial reasoning chain as context, the LVLM continues to generate subsequent reasoning steps, resulting in an extended reasoning chain defined as $S_{rest} = [s_{j+1}', s_{j+2}', \dots, s_{n'}']$. Here, $S_{rest}$ represents the model's attempt to self-correct previously identified erroneous reasoning while continuing to answer the given question $Q$.

We again employ an LLM-based judge to identify the presence of these human corrections within the model-generated steps in $S_{rest}$. The model returns $1$ if any of the reasoning steps within $S_{rest}$ semantically match the corresponding human correction steps from $[s_i, s_{i+1}, \dots, s_j]$, and $0$ otherwise.

Aggregating across all the incorrect subsequences identified by annotators, we compute the score for \textit{Self-Heal} as the fraction of instances in which the LVLM successfully corrects the injected erroneous reasoning step.

\section{Experiment and Results}
\label{sec:05-experiment}
In this section, we perform a series of experiments on multiple state-of-the-art LVLMs to examine their visual grounding abilities and overall reasoning quality. We found that accuracy climbs almost linearly with how model’s focus aligns to relevant regions, and scaling boosts its performance on region focusing, reasoning and self-correction.

\subsection{Experimental Settings}
We evaluate a total of 12 open-source, general-purpose LVLMs without test-time inference scaling capabilities, including LLaVA-OneVision-Chat \citep{xiong2024llavaovchat}, InternVL 2.5-MPO \citep{wang2024mpo}, Qwen2.5-VL \citep{bai2025qwen25vltechnicalreport}, Llama-3.2 \citep{llama3modelcard}, Aya Vision \citep{aya-vision}, and Gemma 3\citep{gemma3}. Additionally, we evaluate the only open-source model equipped with test-time inference scaling to the best of our knowledge, QvQ \citep{qvq-72b-preview}\footnote{\label{fn:qvq}QvQ was evaluated zero-shot because it natively generates extended step-by-step visual reasoning, and its release notes caution that it may gradually lose focus on image content during multi-step visual reasoning. We additionally evaluated QvQ-72B with the identical bounding-box ICL demonstrations used for all other models. \score{} remains nearly unchanged (macro-F1: $60.26 \rightarrow 61.97$; micro-F1: $66.69 \rightarrow 67.08$), while answer accuracy decreases rather than improves (Full: $59.89 \rightarrow 34.67$; Short: $64.04 \rightarrow 36.96$). Thus the zero-shot protocol did not advantage QvQ.}, and two proprietary LVLMs from OpenAI: GPT-4o-mini \citep{gpt4o} and GPT-4.1 \citep{GPT4-1}.

As described in \autoref{sec:04-task_formulation}, for each image-question pair $(I, Q)$ in the test set, models are instructed to generate detailed reasoning steps required to answer the provided question, based on the associated image with clearly marked bounding boxes. As constructed in section \ref{sec:03-dataset-task-input}, the test set includes both relevant and adversarial boxes by default, and models are evaluated under this mixed candidate set. Both the reasoning steps and final answers produced by each model are extracted for evaluation. Additional experimental details can be found in the \autoref{sec:method-detail}.



\begin{table*}[!htpb]
\centering
\resizebox{\textwidth}{!}{%
\begin{tabular}{lcccccccccc}
\hline
\multirow{2}{*}{Model} &
  \multirow{2}{*}{Size} &
  \multicolumn{1}{c|}{\multirow{2}{*}{\begin{tabular}[c]{@{}c@{}}Think \\ Capability\end{tabular}}} &
  \multicolumn{2}{c|}{Final Answer} &
  \multicolumn{3}{c|}{\score{} (Macro)} &
  \multicolumn{3}{c}{\score{} (Micro)} \\
 &
   &
  \multicolumn{1}{c|}{} &
  Full &
  \multicolumn{1}{c|}{Short} &
  Precision &
  Recall &
  \multicolumn{1}{c|}{F1} &
  Precision &
  Recall &
  F1 \\ \hline
\multicolumn{11}{c}{\textit{Small ($\sim$7B) Open LVLM}} \\ \hline
LLaVA OneVision &
  7B &
  \multicolumn{1}{c|}{No} &
  49.00 &
  \multicolumn{1}{c|}{54.01} &
  48.31 &
  35.33 &
  \multicolumn{1}{c|}{37.79} &
  77.78 &
  32.72 &
  46.06 \\
InternVL 2.5-MPO &
  8B &
  \multicolumn{1}{c|}{No} &
  49.86 &
  \multicolumn{1}{c|}{50.00} &
  65.61 &
  53.50 &
  \multicolumn{1}{c|}{54.12} &
  66.61 &
  50.90 &
  57.71 \\
Aya Vision &
  8B &
  \multicolumn{1}{c|}{No} &
  45.13 &
  \multicolumn{1}{c|}{50.00} &
  78.63 &
  50.32 &
  \multicolumn{1}{c|}{56.74} &
  89.02 &
  43.83 &
  58.74 \\
InternVL 3 &
  9B &
  \multicolumn{1}{c|}{No} &
  49.43 &
  \multicolumn{1}{c|}{\textbf{56.45}} &
  69.62 &
  58.09 &
  \multicolumn{1}{c|}{\textbf{58.31}} &
  66.60 &
  53.34 &
  \textbf{59.24} \\ 
Qwen2.5-VL &
  7B &
  \multicolumn{1}{c|}{No} &
  \textbf{51.00} &
  \multicolumn{1}{c|}{55.30} &
  53.49 &
  42.60 &
  \multicolumn{1}{c|}{43.15} &
  67.93 &
  38.51 &
  49.15 \\ \hline
\multicolumn{11}{c}{\textit{Medium (11$\sim$32B) Open LVLM}} \\ \hline
Gemma 3 &
  12B &
  \multicolumn{1}{c|}{No} &
  50.00 &
  \multicolumn{1}{c|}{54.58} &
  76.49 &
  69.15 &
  \multicolumn{1}{c|}{\textbf{67.69}} &
  73.03 &
  64.45 &
  \textbf{68.47} \\
InternVL 2.5-MPO &
  26B &
  \multicolumn{1}{c|}{No} &
  36.39 &
  \multicolumn{1}{c|}{40.69} &
  71.47 &
  58.17 &
  \multicolumn{1}{c|}{59.44} &
  76.59 &
  53.30 &
  62.86 \\
Gemma 3 &
  27B &
  \multicolumn{1}{c|}{No} &
  50.29 &
  \multicolumn{1}{c|}{52.44} &
  69.46 &
  73.23 &
  \multicolumn{1}{c|}{66.48} &
  66.61 &
  70.07 &
  68.30 \\
Aya Vision &
  32B &
  \multicolumn{1}{c|}{No} &
  47.71 &
  \multicolumn{1}{c|}{52.44} &
  84.68 &
  57.23 &
  \multicolumn{1}{c|}{63.15} &
  87.74 &
  51.24 &
  64.70 \\
Qwen2.5-VL &
  32B &
  \multicolumn{1}{c|}{No} &
  \textbf{59.74} &
  \multicolumn{1}{c|}{\textbf{63.32}} &
  73.85 &
  57.03 &
  \multicolumn{1}{c|}{59.18} &
  69.07 &
  51.42 &
  58.95 \\ \hline
\multicolumn{11}{c}{\textit{Large (72$\sim$90B) Open LVLM}} \\ \hline
Llama 3.2 Vision &
  90B &
  \multicolumn{1}{c|}{No} &
  35.67 &
  \multicolumn{1}{c|}{39.97} &
  46.28 &
  37.55 &
  \multicolumn{1}{c|}{38.29} &
  83.35 &
  33.49 &
  47.78 \\ 
Qwen2.5-VL &
  72B &
  \multicolumn{1}{c|}{No} &
  \textbf{61.03} &
  \multicolumn{1}{c|}{\textbf{66.05}} &
  60.80 &
  80.95 &
  \multicolumn{1}{c|}{\textbf{63.37}} &
  56.82 &
  75.43 &
  64.81 \\
QvQ &
  72B &
  \multicolumn{1}{c|}{Yes} &
  59.89 &
  \multicolumn{1}{c|}{64.04} &
  47.43 &
  89.35 &
  \multicolumn{1}{c|}{60.26} &
  53.51 &
  88.46 &
  \textbf{66.69} \\ \hline
\multicolumn{11}{c}{\textit{Proprietary LVLMs}} \\ \hline
GPT 4o mini &
  N/A &
  \multicolumn{1}{c|}{No} &
  51.72 &
  \multicolumn{1}{c|}{55.01} &
  69.03 &
  79.14 &
  \multicolumn{1}{c|}{\textbf{66.77}} &
  63.88 &
  67.71 &
  65.74 \\
GPT 4.1 &
  N/A &
  \multicolumn{1}{c|}{No} &
  \textbf{71.78} &
  \multicolumn{1}{c|}{\textbf{74.36}} &
  65.98 &
  71.10 &
  \multicolumn{1}{c|}{63.73} &
  61.62 &
  74.01 &
  \textbf{67.25} \\ \hline
\end{tabular}%
}
\vspace{-0.8em}
\caption{Performance comparison. For QvQ, we report the zero-shot setting; with standardized ICL, \score{} is stable while answer accuracy drops, as detailed in footnote~\ref{fn:qvq}. The best performing models for each metric are shown in bold.}
\label{tab:main-result}
\vspace{-1em}
\end{table*}

\subsection{Main Results}

\paragraph{Focusing on the right place leads to higher answer performance.} Across the 15 LVLMs evaluated in \autoref{tab:main-result}, we find that the models with higher \score{} tend to perform better in answering the question itself. This pattern holds for both small- and large-scales, suggesting that the ability to selectively attend to relevant regions is a strong indicator for the correctness of models' final answer. For example, {Gemma 3 12B}, which scored $67.7\%$ in attention, achieved $54.6\%$ in short answer accuracy, while {Llama 3.2-90B}, scored $38.3\%$ in attention only and subsequently achieved $39.97\%$ in answer accuracy. We notice that the {Qwen2.5-VL} models generally tend to have lower attention scores than their peers despite high short-answer accuracies. We speculate that the attention scores gap with other models probably comes from Qwen's training recipe \citep{bai2025qwen25vltechnicalreport}, which includes pre-training on various VQA datasets, and grounding data with absolute position references, which may interfere with the models' ability to make \emph{relative} references using specific schemes such as our bounding boxes. 

\begin{figure}[!htbp]
\centering
\includegraphics[width=\columnwidth]{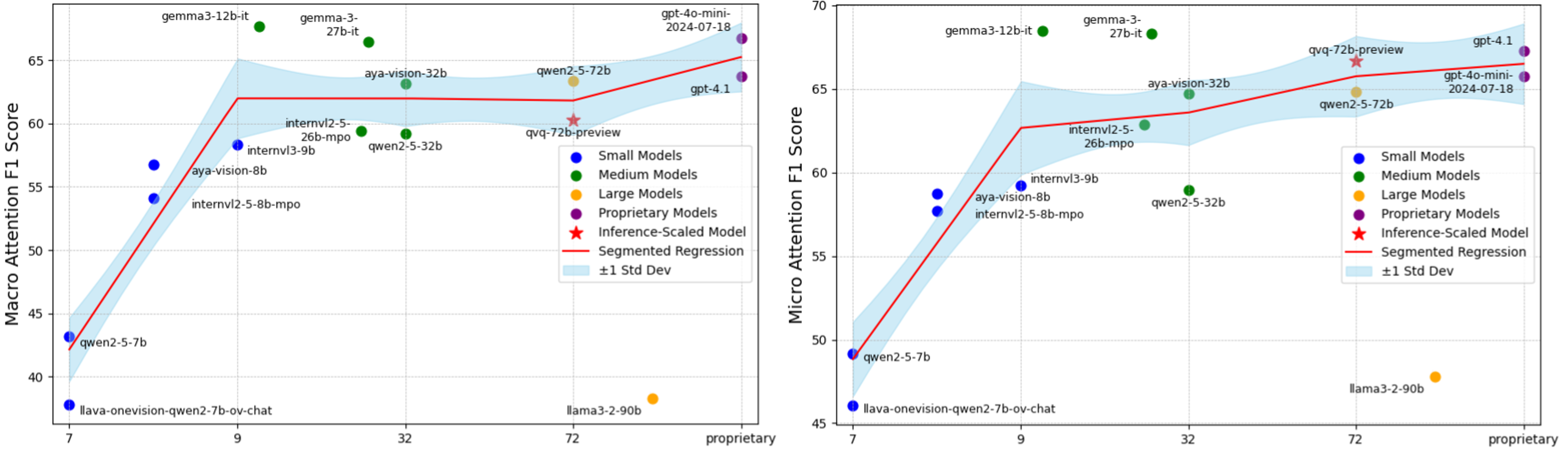}
\vspace{-1.5em}
\caption{Scaling curve for \score{}}
\label{fig:size-to-f1}
\vspace{-2em}
\end{figure}

\paragraph{Region focus as an emergent ability of larger models.} As shown in \autoref{fig:size-to-f1}, we observe that scaling model sizes could lead to better performance in utilizing the relevant regions. Medium-sized (11-32B) LVLMs are on the sweet spot in terms of both region attention performance and final answer performance, as we can see that they generally tend to have higher region attention scores than smaller models. This is followed by higher final answer scores, in terms of both full and short answers. Large-sized (72B+) models seem to provide further performance gains over medium-sized models. For {Qwen2.5-VL} models (7B, 32B, 72B) with lower attention scores than their respective peers, we can still see that the overall trend still holds within these three models.

\paragraph{Attention capabilities of test-time scaling models.} It is interesting to note that the only vision test-time scaling model we test, {QvQ}, achieves comparable performance to {Qwen2.5-VL} 72B, despite not being provided with our in-context learning examples of utilizing bounding boxes for reasoning. We can see that as part of its test-time scaling, {QvQ} could already generate reasoning that effectively considers the bounding boxes we inject into test images. However, the model tends to perform exhaustive coverage of all bounding boxes, as evidenced by low precision in micro and macro attention scores. While it remains to be seen whether this would be the case for other comparable models, we see a great potential of better visual grounding to be achieved with test-time scaling.


\begin{table}[!htpb]
\centering
\vspace{-0.8em}
\begin{tabular}{lc|cc}
\hline
\multirow{2}{*}{Model} & \multicolumn{1}{l|}{\multirow{2}{*}{Size}} & \multicolumn{2}{c}{Accuracy} \\
             & \multicolumn{1}{l|}{} & \textit{StepSense}  & Answer \\ \hline
QwenVL 2.5   & 7B                    & 61.38 & 69.98       \\
InternVL 2.5 & 8B                    & 48.99 & 48.69       \\
Gemma 3      & 27B                   & 52.13 & 55.1        \\
GPT 4o mini  & N/A                   & 53.89 & 59.26       \\ \hline
\end{tabular}
\vspace{-0.8em}
\caption{\textit{StepSense} and final answer accuracy}
\label{tab:annotation-perf}
\vspace{-1em}
\end{table}

\begin{table}[!htpb]
\centering
\begin{tabular}{lc|c}
\\ \hline
Model           & Size & \textit{Self-Heal} \\ \hline
LLaVA OneVision & 7B   & 25.99    \\
Gemma 3         & 12B  & 31.5     \\
Qwen2.5-VL      & 7B   & 35.03    \\
Qwen2.5-VL      & 32B  & 41.38    \\
Qwen2.5-VL      & 72B  & 46.89   \\ \hline
\end{tabular}
\vspace{-0.8em}
\captionof{table}{\textit{Self-Heal} accuracy}
\label{tab:correction-performance}
\vspace{-2em}
\end{table}

\paragraph{Sound intermediate reasoning, sound final answer.} Human-annotated evaluation from \autoref{tab:annotation-perf} shows that models with higher sentence-level reasoning accuracy almost always enjoy higher final-answer accuracy - for example, {Qwen2.5-VL} 7B leads both metrics ($61.4\%$ vs. $70.0\%$), while {InternVL} 8B trails on both. Together with our findings in region focus, this supports a three-step chain: a precise perception with reliable reasoning can result in correct answers. Improving any links in that chain, and especially the visual grounding link, therefore remains a direct path to better overall performance.

\paragraph{Stronger models correct themselves better.} The self-correction results in \autoref{tab:correction-performance} echo the same scaling story.  Each jump in {Qwen2.5-VL} size yields roughly a six-point boost in correction accuracy. We also tested two models, LLaVA-OneVision 7B and Gemma 3 12B, which achieved short answer performance similar to {Qwen2.5-VL} 7B. However, we can see that both models lag noticeably behind {Qwen2.5-VL} 7B, showing that stronger model's tend to have better self-correction capabilities. as well. 
\subsection{Error Analysis}

We randomly sample 100 instances of failed reasoning from all 15 LVLMs we have tested and analyze their failure points. \autoref{fig:error_analysis} shows a list of failed examples. We identify the following common error patterns:

\begin{figure}[!htbp]
\vspace{-1em}
\centering
\includegraphics[width=\columnwidth]{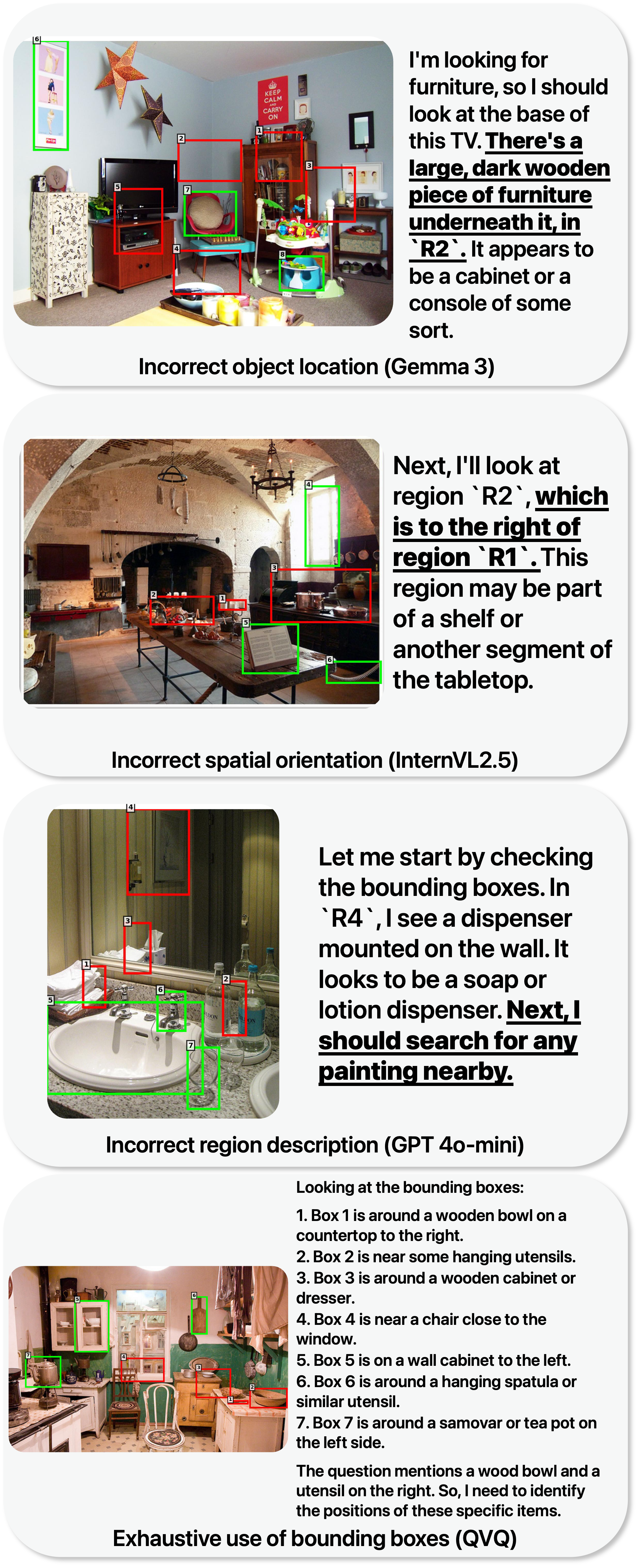}
\vspace{-2em}
\caption{\textbf{Examples of failed reasoning.} Adversarial boxes are labeled in green.}
\label{fig:error_analysis}
\vspace{-1.5em}
\end{figure}

\paragraph{Exhaustive coverage of all regions} In many cases, the models tend to exhaustively consider all bounding boxes that appear in the image, despite our instruction to selectively utilize bounding boxes that are relevant for answering the question. Moreover, it is often found that models decide to continue iterating through every single bounding boxes, even though it has already inspected other bounding boxes and realized the answer to the question. This is a potentially problematic behavior as it is not only inefficient to consider irrelevant regions, but also could increase the chances of hallucination over analysis into regions that are actually relevant. We find that this is especially the case for QvQ, which is trained to generate longer chains of reasoning by default.


\paragraph{Incorrect object location} Another pattern we have identified is that even though the models correctly identified relevant details required to answer the question, they often hallucinate by claiming that such details appear in unrelated bounding boxes. For example, in the top left example of \autoref{fig:error_analysis}, it is true that there is a dark wood piece of furniture under the TV.  However, the model incorrectly claims that it is located in R2, not R5. 

\paragraph{Wrong spatial relations between the bounding boxes} Related to the issue of incorrect object location, we also see that the models often seem to experience issues with spatial orientation, as evidenced by the bottom left image of \autoref{fig:error_analysis}. While R2 is actually located to the left of region R1, the model incorrectly claims that it is actually to the \emph{right} of R1.

\paragraph{Incorrect region description} Even when the model has identified and analyzed relevant bounding boxes, we found that the model output incorrect details of the region. In the bottom right image of \autoref{fig:error_analysis}, the model was able to correctly choose R4 as part of its analysis and identifies the soap dispenser in the region. However, it fails to see that the painting in question is also in R4.

\section{Conclusion}
\label{sec:06-conclusion}
This paper introduced \model{}, a novel benchmark designed to evaluate grounded visual reasoning over natural scenes. We evaluated 15 LVLMs ranging from 7B to 70B+ parameters across four dimensions: final answer correctness, reasoning validity, grounding fidelity, and self-correction ability. Our analysis shows that the models that could focus better on relevant regions tend to achieve better task performance, and such focusing ability improves with increased model sizes. However, we also found that current SOTA LVLMs all suffer from various types of region focus failures, including incorrect object location, wrong spatial relations, and inaccurate region descriptions.

\section*{Limitation}
\model{} is scoped to grounded visual reasoning over natural scenes. It builds on GQA and AiR-D for aligned questions, scene-graph annotations, and human eye-tracking data, but consequently does not cover OCR, document, chart, mathematical-diagram, or medical-image reasoning. Answer accuracy and Self-Heal rely on LLM-based semantic-equivalence judgments. We validate these judges in \autoref{sec:app-judge-validation} against multiple LLMs and human labels, though LLM-judge dependence remains a limitation; StepSense is fully human-annotated. Public GQA text could at most inflate final-answer accuracy and cannot affect \score{}, StepSense, or Self-Heal, whose targets were newly created for \model{}. We observe no ceiling effect: even the best models show substantial room for improvement in both answer accuracy and process scores.

\section*{Acknowledgment}
The work was supported by NSF CAREER award IIS2048001 and the DARPA CCU program. Contents do not necessarily represent the official views of, nor an endorsement by, DARPA, or the US Government.

We would acknowledge the help from Yuxin Li, William Huang from University of Pennsylvania for the annotation.

\bibliography{reference}


\clearpage
\appendix
\section{Data Statistics}
\begin{figure*}[p]
    \centering
    \includegraphics[width=\textwidth]{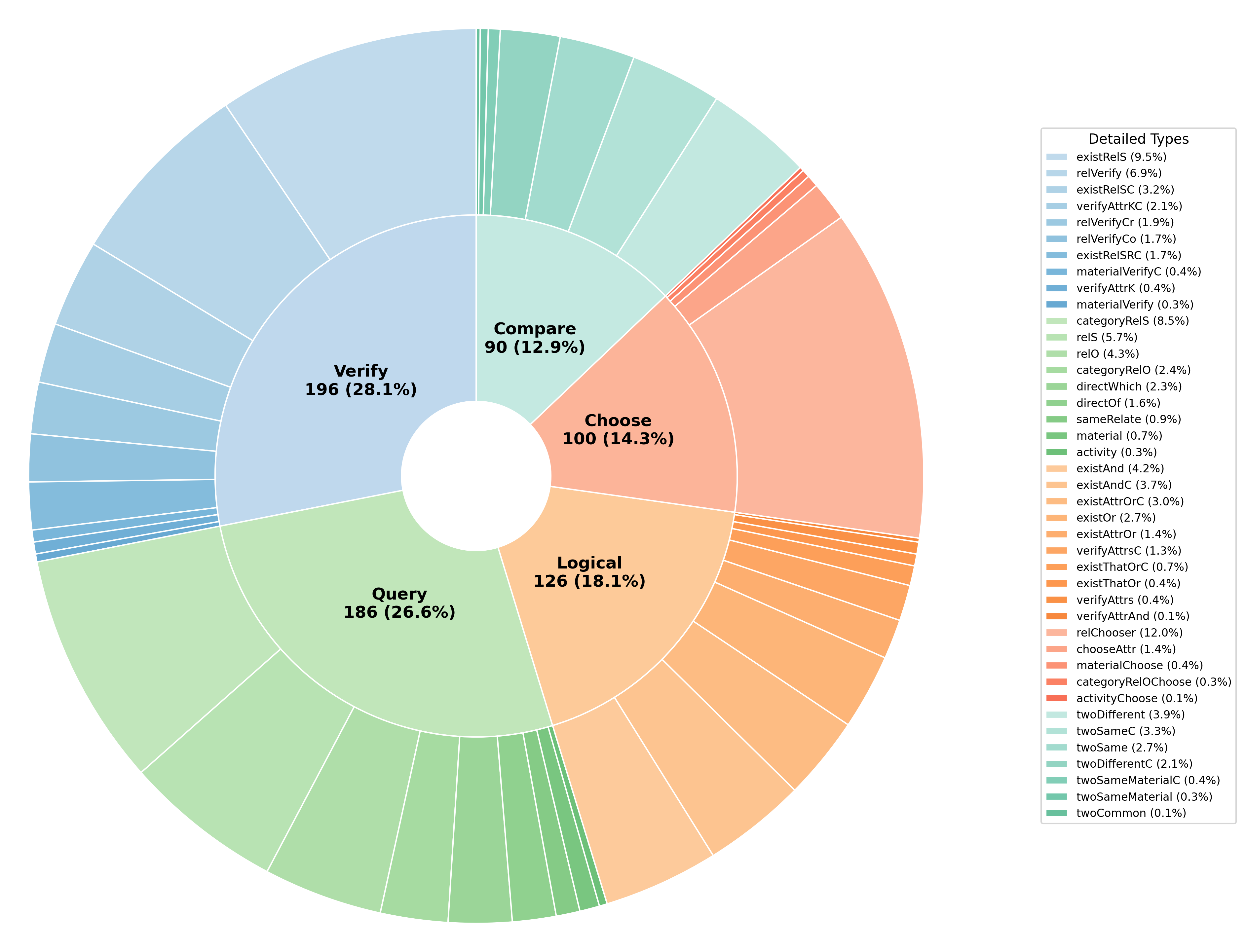}
    \caption{Detailed question type distributions for \model{} benchmark.}
    \label{fig:type-distribution-full}
\end{figure*}

Detailed question type distributions for \model{} benchmark is in \autoref{fig:type-distribution-full}. The data statistics is in \autoref{tab:data-stat}.

\begin{figure}[!htbp]
    \centering
    \begin{tabular}{l|ccc}
    \hline
    \textbf{Set}       & \textbf{WS*} & \textbf{Dev} & \textbf{Test} \\ \hline
    \textbf{\# of Instances}       & 5534         & 198          & 698           \\
    \textbf{\# of Steps} & 69523        & 2373         & 8403          \\
    Avg. Words/Step         & 23           & 18           & 19            \\
    Avg. Steps/Inst.         & 12.00        & 11.98        & 12.03         \\
    \textbf{\# of Correction}      & N/A          & 641          & 2700          \\
    Avg. Words         & N/A          & 18           & 19            \\ \hline
    \end{tabular}
    \captionof{table}{Dataset statistics.\\\textit{WS: Weakly supervised split}}
    \label{tab:data-stat}
\end{figure}

\section{\score{} Calculation Example} \label{sec:app-score-example}
\begin{figure}[!htbp]
    \centering
    \includegraphics[width=\columnwidth]{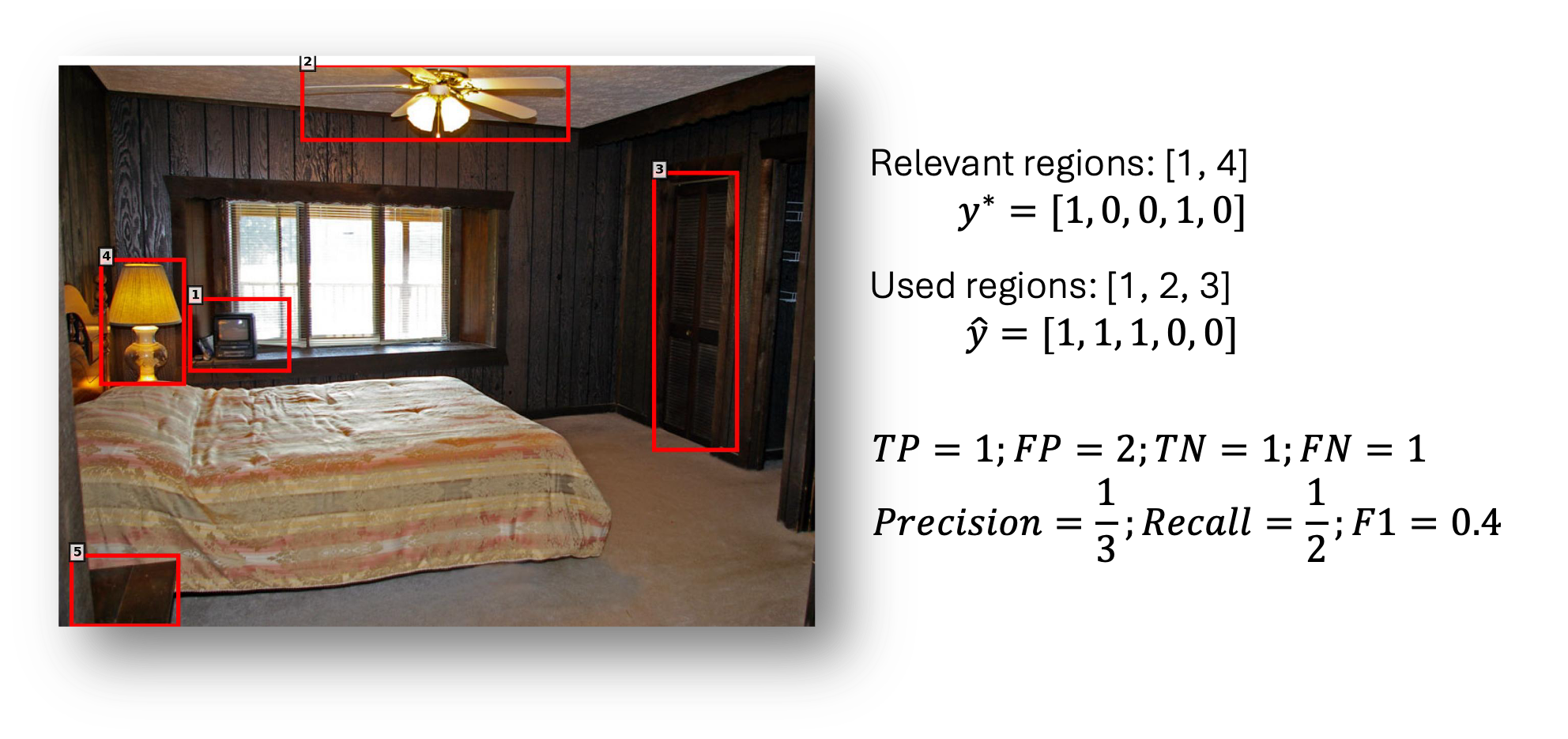}
    \caption{Example on how \score{} are calculated.}
    \label{fig:magiscore-calculation}
\end{figure}

The example can be found in \autoref{fig:magiscore-calculation}.

\section{Data Examples} \label{sec:app-data-example}
\begin{figure*}[!tbp]
    \centering
    \includegraphics[width=\textwidth]{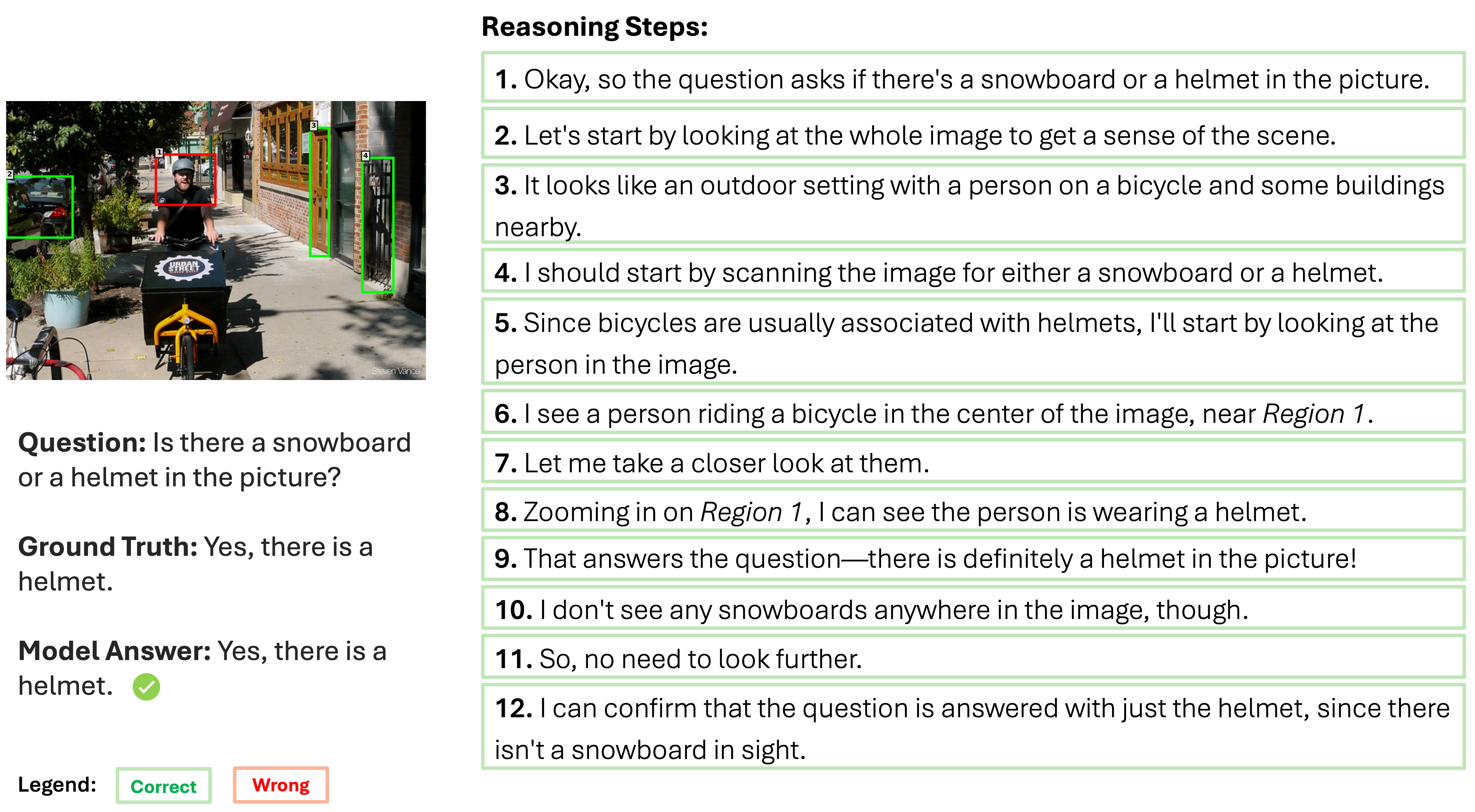}
    \caption{Example where both reasoning and answer are correct}
    \label{fig:example-perfect}
\end{figure*}

\begin{figure*}[!tbp]
    \centering
    \includegraphics[width=\textwidth]{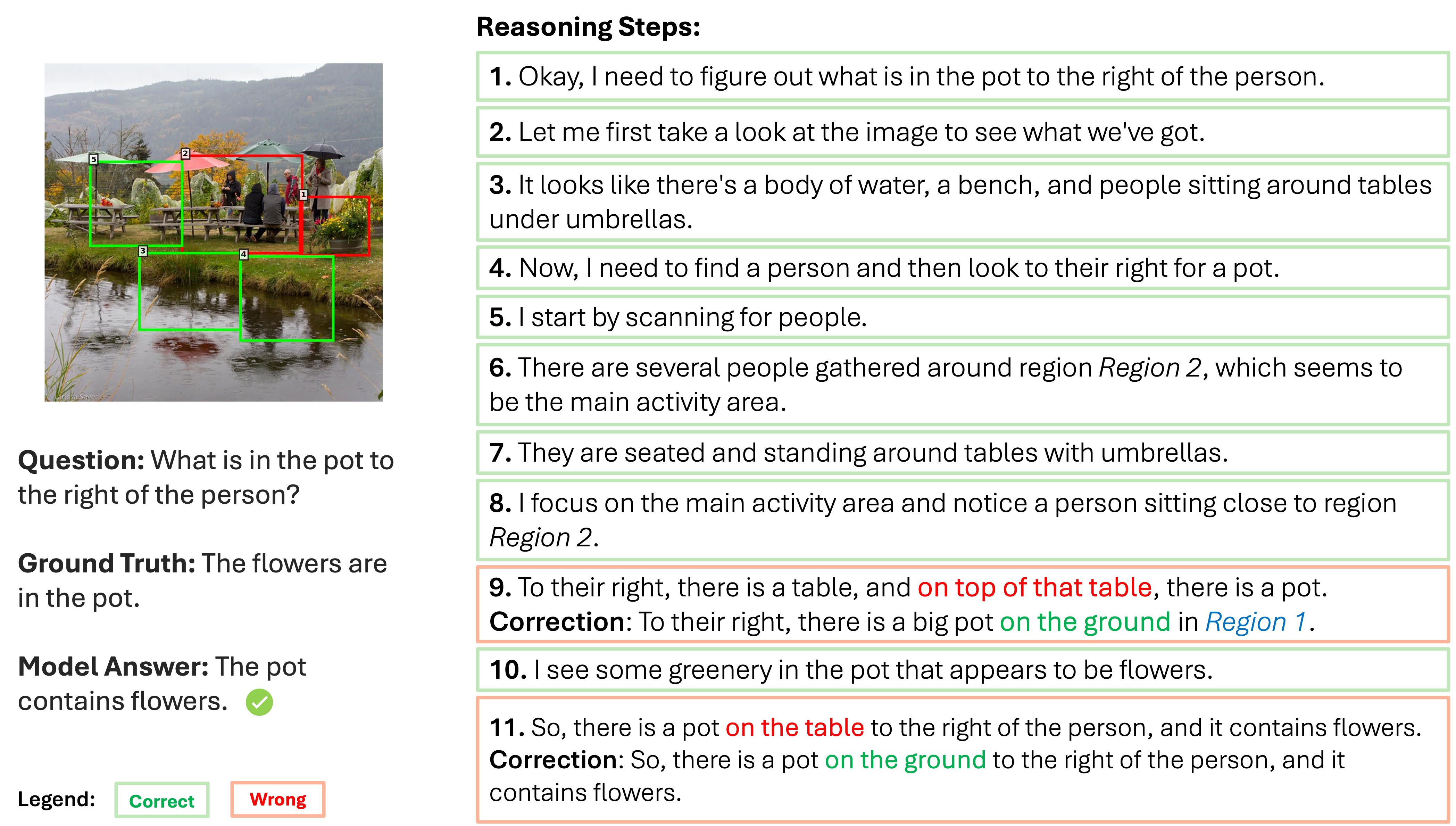}
    \caption{Example where reasoning is partially incorrect and answer are correct}
    \label{fig:example-wrong-reason-right-answer}
\end{figure*}

\begin{figure*}[!tbp]
    \centering
    \includegraphics[width=\textwidth]{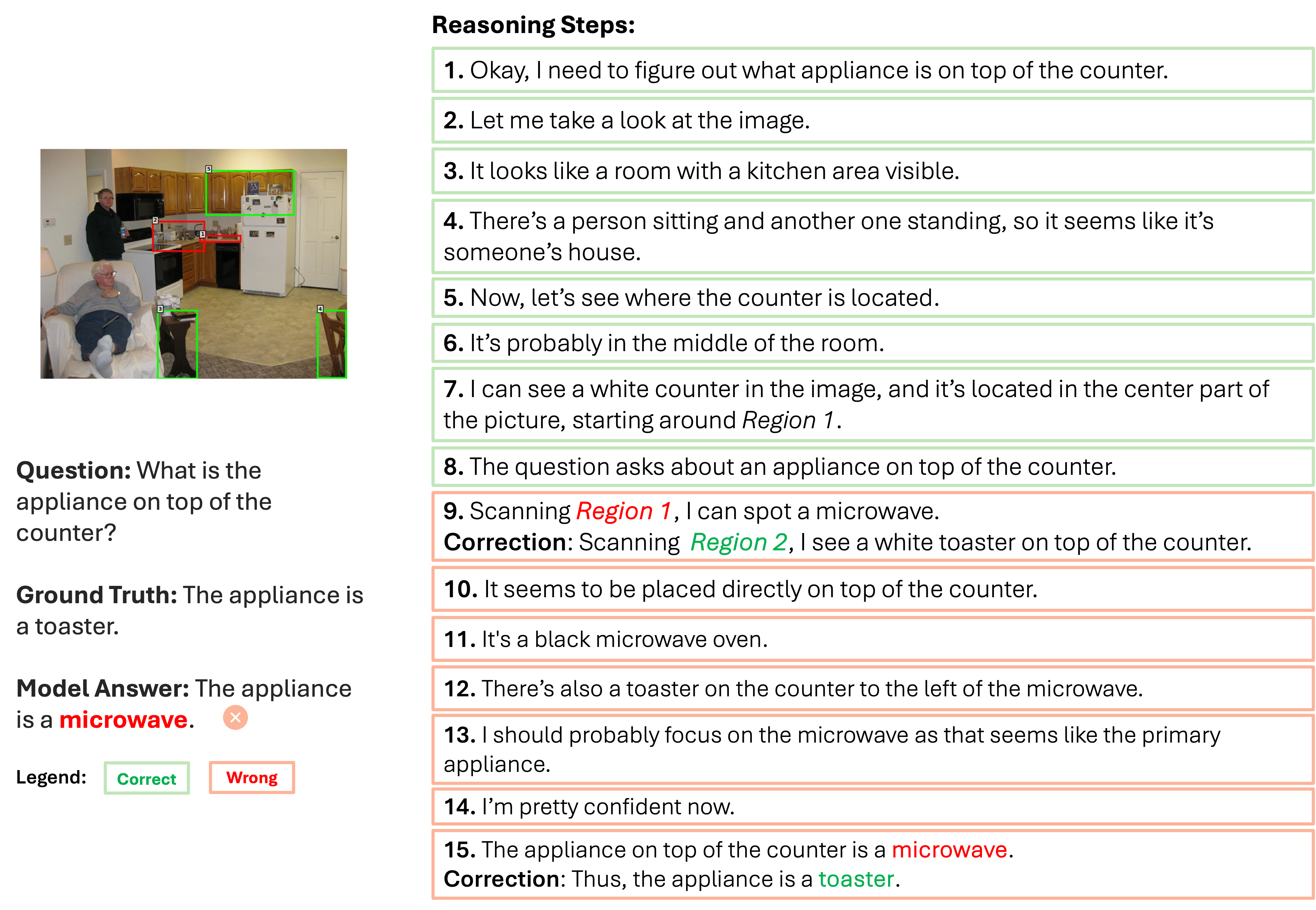}
    \caption{Example where both reasoning and answer are incorrect, red boxes without correction means manual steer of reasoning from annotator.}
    \label{fig:example-bad}
\end{figure*}

\begin{figure*}[!tbp]
    \centering
    \includegraphics[width=\textwidth]{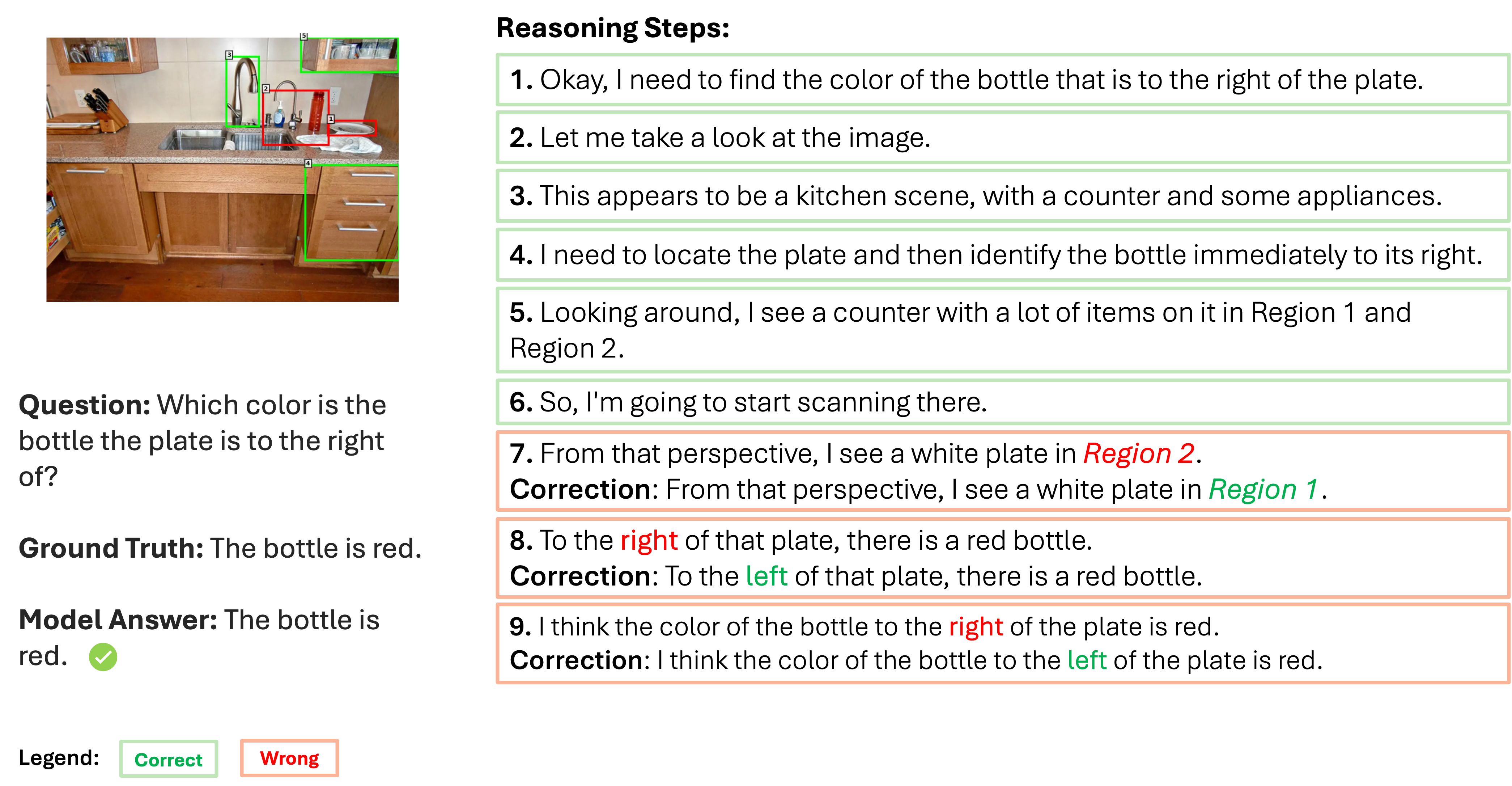}
    \caption{Example where reasoning is partially incorrect and answer are correct}
    \label{fig:example-wrong-reason-right-answer-2}
\end{figure*}

Different examples can be found in \autoref{fig:example-perfect}, \autoref{fig:example-wrong-reason-right-answer}, \autoref{fig:example-bad}, \autoref{fig:example-wrong-reason-right-answer-2}.

\section{Experimental Details} \label{sec:method-detail}
We evaluate 13 open LVLMs and 2 proprietary LVLMs as shown in \autoref{tab:main-result}. We use \texttt{vllm} on all models except \texttt{LLaVA}, \texttt{Aya Vision}, and \texttt{InternVL} which uses \texttt{transformers} library. We set the sampling temperature with the default temperature provided with the each model for the three main tasks. All experiments are performed on either a single Nvidia A100-80GB GPU, Nvidia A100-40GB GPU, or Nvidia H100-80GB GPU.

LLM-assisted evaluation uses \texttt{Qwen/Qwen2.5-72B-Instruct-AWQ} with greedy decoding \texttt{(temp=0)} using \texttt{vllm} on a single Nvidia A100-SXM-80GB GPU.


\section{Details on Task-Input Processing}
Algorithm can be found in \autoref{alg:pre-process}.

\section{Cross-Judge Agreement and Human Validation} \label{sec:app-judge-validation}

We also graded the evaluated models using \texttt{Llama-3.3-70B-Instruct} and \texttt{Gemma-3-27B-IT} in addition to \texttt{Qwen2.5-72B}. Mean pairwise agreement is 94.2\%, individual $\kappa$ values range from 0.79--0.95, and ranking correlations are $\rho = 0.86$--$0.95$. All three judges identify the same top two and bottom two models; changes are confined to mid-ranked models separated by less than four accuracy points. Relative to the other two judges' mean, the Qwen judge is $+0.8$ accuracy points on \texttt{Qwen2.5-VL} outputs and $+0.7$ points on the other nine models, indicating no Qwen-family-specific leniency. On a randomly sampled subset spanning one model per size class, human annotation with the same yes/no protocol matched the judge on 100\% of decisions. At full scale, the Qwen2.5-72B judge reproduces the 698 human final-answer labels with 91.5\% agreement ($\kappa = 0.83$). Please refer to \autoref{tab:cross-judge} for more details.

\section{Model Trained with Weakly Supervised Split} \label{sec:app-weak-supervision}
We conduct training using the weakly supervised data. We use \texttt{Qwen2.5-VL-7B} as the base checkpoint. The trained model improves final answer (full) from 51.0 to 58.1 and \score{} (macro) F1 from 43.2 to 82.2.

\section{Capability Comparison with Prior Benchmarks}

\autoref{tab:capability} provides a supplementary comparison of \model{} with prior benchmarks, covering data characteristics, annotation types, and evaluation protocols.

\onecolumn
\begin{algorithm}[!hbtp]
\caption{Generate and Save Relevant Bounding Boxes}
\label{alg:pre-process}
\DontPrintSemicolon

\SetKwFunction{FFindBboxes}{FindBboxes}
\SetKwFunction{FRemoveOverlap}{RemoveOverlap}
\SetKwFunction{FAdvBoxes}{AdvBoxes}
\SetKwProg{Fn}{Function}{:}{\KwRet}

\Fn{\FFindBboxes{$s\_path$, $img\_size$}}{
  $M \gets \text{load\_gray}(s\_path)\rightarrow\text{resize}(img\_size)\rightarrow[0,1]$ \;
  $B \gets \bigl(M > \text{thresh}\bigr)$ \tcp*{binarize}
  $B \gets \text{morph\_clean}(B)$ \;
  \KwRet all $\text{bbox}(r)$ for each connected region $r$ with $\text{area}(r)\ge area_{min}$\;
}

\Fn{\FRemoveOverlap{$\mathcal{B}$}}{
  \KwRet $\mathcal{B}$ pruned by IoU / coverage / temporal rule at threshold\;
}

\Fn{\FAdvBoxes{$id$, $\mathcal{G}$, $S$}}{
  $\mathcal{C} \gets \text{cached\_adv}(id)$ \;
  $\mathcal{C} \gets $\FRemoveOverlap{$\mathcal{C} \cup \mathcal{G}$} \;
  $\mathcal{C} \gets \text{size\_filter}(\mathcal{C})$ \;
  \KwRet sample or synthesize boxes so that $|\mathcal{C}|$ meets the required quota\;
}
\;

\ForEach{question $(qid, q)$ in \texttt{questions}}{
  $I \gets \text{read\_image}(q.\text{imageId})$ \;
  $\mathcal{S} \gets \varnothing$ \;

  \For{$t \in \{0,1,2\}$}{
    $m \gets \text{load\_saliency\_map}(qid,\,t)$ \;
    $\mathcal{S} \gets \mathcal{S} \cup $\FFindBboxes{$m,\,I.\text{size}$} \;
  }
  \texttt{label} $\mathcal{S}$ as \textit{non\_adv}\;

  $\mathcal{G} \gets$ ground--truth objects (gqa objects)\;
  $\mathcal{B} \gets \text{merge}(\mathcal{G},\,\mathcal{S})$\;
  $\mathcal{B} \gets $\FRemoveOverlap{$\mathcal{B},\text{coverage},\text{user\_thr}$}\;

  $\mathcal{B} \gets \mathcal{B} \cup $\FAdvBoxes{$q.\text{imageId},\mathcal{B},I.\text{size}$}\;

  \texttt{draw\&save}$(I,\,\mathcal{B})$\;
}
\end{algorithm}
\twocolumn

\begin{table*}[!htbp]
\centering
\small
\resizebox{0.8\textwidth}{!}{%
\begin{tabular}{lccc}
\toprule
Output model & Qwen $\leftrightarrow$ Llama & Qwen $\leftrightarrow$ Gemma & Llama $\leftrightarrow$ Gemma \\
\midrule
Aya-Vision-8B & 95.0 (0.90) & 94.8 (0.90) & 96.1 (0.92) \\
InternVL2.5-8B-MPO & 94.3 (0.89) & 91.7 (0.83) & 95.4 (0.91) \\
InternVL3-9B & 94.1 (0.88) & 92.7 (0.85) & 94.3 (0.88) \\
LLaVA-OneVision-7B & 92.7 (0.85) & 91.0 (0.82) & 95.8 (0.92) \\
Qwen2.5-VL-7B & 94.6 (0.89) & 93.4 (0.87) & 94.8 (0.90) \\
Aya-Vision-32B & 93.8 (0.88) & 93.3 (0.86) & 96.0 (0.92) \\
Gemma-3-12B-IT & 96.0 (0.92) & 94.6 (0.89) & 97.4 (0.95) \\
Gemma-3-27B-IT & 96.4 (0.93) & 96.4 (0.93) & 97.7 (0.95) \\
InternVL2.5-26B-MPO & 90.8 (0.80) & 90.4 (0.79) & 93.3 (0.85) \\
Qwen2.5-VL-32B & 96.0 (0.92) & 93.8 (0.87) & 94.7 (0.89) \\
Qwen2.5-VL-72B & 95.6 (0.91) & 93.3 (0.86) & 91.7 (0.83) \\
GPT-4o-mini & 94.6 (0.89) & 92.4 (0.85) & 93.6 (0.87) \\
\midrule
Mean & 94.5 (0.89) & 93.1 (0.86) & 95.1 (0.90) \\
\bottomrule
\end{tabular}%
}
\caption{Cross-judge agreement by output model. Each cell reports percent agreement (Cohen's $\kappa$).}
\label{tab:cross-judge}
\end{table*}

\begin{table}[!htbp]
\centering
\small
\resizebox{\columnwidth}{!}{%
\begin{tabular}{lc}
\toprule
Output model & Human $\leftrightarrow$ Qwen2.5-72B \\
\midrule
InternVL2.5-8B-MPO & 100.0 (1.00) \\
Gemma-3-12B-IT & 100.0 (1.00) \\
GPT-4o-mini & 100.0 (1.00) \\
Qwen2.5-VL-72B & 100.0 (1.00) \\
\midrule
Mean & 100.0 (1.00) \\
\bottomrule
\end{tabular}%
}
\caption{Human agreement with the Qwen2.5-72B judge on a sampled audit.}
\label{tab:human-judge}
\end{table}

\begin{table*}[!htb]
\centering
\small
\resizebox{\textwidth}{!}{%
\begin{tabular}{lccccccc}
\toprule
Dataset & Natural images & Open-ended QA & Region annotations & Relevant \& adversarial regions & Step-level annotation & Self-correction eval. & Semantic answer judging \\
\midrule
GQA & \checkmark & \checkmark & \checkmark & -- & -- & -- & -- \\
Visual Genome & \checkmark & \checkmark & \checkmark & -- & -- & -- & -- \\
Visual7W & \checkmark & -- & \checkmark & -- & -- & -- & -- \\
RefCOCO/g & \checkmark & -- & \checkmark & -- & -- & -- & -- \\
Flickr30k Entities & \checkmark & -- & \checkmark & -- & -- & -- & -- \\
VCR & \checkmark & -- & \checkmark & -- & -- & -- & -- \\
CLEVR & -- & \checkmark & \checkmark & -- & -- & -- & -- \\
Visual CoT & \checkmark & \checkmark & \checkmark & -- & -- & -- & -- \\
\model{} & \checkmark & \checkmark & \checkmark & \checkmark & \checkmark & \checkmark & \checkmark \\
\bottomrule
\end{tabular}%
}
\caption{Capability comparison with prior benchmarks. \model{} is the only benchmark combining all seven capabilities. Visual CoT differs on several dimensions central to \model{}: it uses a single evidence box per question; grounding is scored as box detection at IoU $\geq 0.5$; reasoning steps are largely GPT-generated, with no human step-level labels on model reasoning; and it does not include a self-correction protocol.}
\label{tab:capability}
\end{table*}

\newpage

\begin{figure*}[!htbp]
    \centering
    \includegraphics[width=395pt]{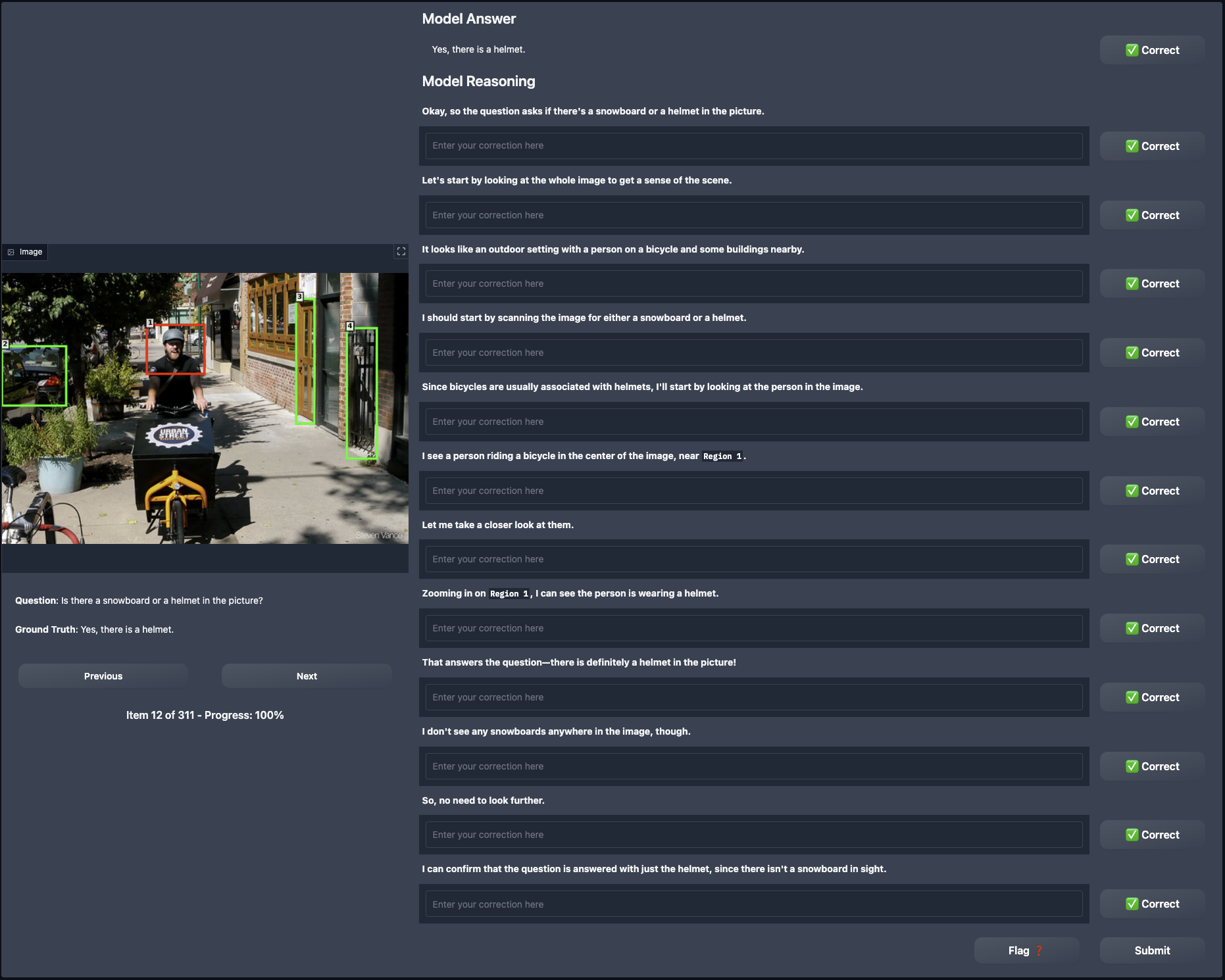}
    \caption{Annotation User Interface}
    \label{fig:apdx-ui}
\end{figure*}

\begin{figure}
\centering
\fontsize{7}{9}\selectfont
\begin{tabularx}{1.0\columnwidth}{X}
\hline
\texttt{\color{cyan}**In-Context Learning examples go here**}

Similarly to the examples provided above, please imagine this person's thought process behind the areas of the image the person looked at in order to solve the question, which are marked with the red bounding boxes.

The thought process should meet the following conditions:\\
- It should describe the broader setting of the image and how that context informs the search for specific details.\\
- It should explain how the bounding boxes (e.g., `R1`, `R2`) were used to narrow down the areas of interest. For instance, it should mention how the image's orientation or the placement of objects guided the focus to certain regions.\\
- It should only consider the bounding boxes that does appear in the image and not refer to any bounding boxes that actually do not exist. Note that all bounding boxes are numbered sequentially, so if we have 4 bounding boxes in the image, there are exactly the following bounding boxes: `R1`, `R2`, `R3`, and `R4`.\\
- It should list specific visual cues noticed within each highlighted region and how these observations lead to identifying the relevant subject (e.g., a man with a backpack).\\
- It should clearly connect each observation to the final answer. It should explain how, based on the evidence gathered from the image, the conclusion was reached (e.g., identifying the man holding a cell phone).\\
\\
\\
Write as if you are this person, in a similar style as above examples.
\\
\\
\#\# Image \texttt{\{im\_num\}} (\texttt{\{num\_bbox\}} bounding box(es))

**Question**: \texttt{\{question\}}\\
**Thought Process**: \\
**Answer**: \\

\hline
\end{tabularx}
\vspace{-1em}
\caption{Prompt for reasoning and answer generation.}
\label{fig:prompt_reasoning}
\end{figure}

\begin{figure}[!htpb]
\centering
\fontsize{7}{9}\selectfont
\begin{tabularx}{1.0\columnwidth}{X}
\hline
\texttt{\color{cyan}**In-Context Learning examples go here**}

Similarly to the examples provided above, please imagine this person's thought process behind the areas of the image the person looked at in order to solve the question, which are marked with the red bounding boxes.

The thought process should meet the following conditions:\\
- It should describe the broader setting of the image and how that context informs the search for specific details.\\
- It should explain how the bounding boxes (e.g., `R1`, `R2`) were used to narrow down the areas of interest. For instance, it should mention how the image's orientation or the placement of objects guided the focus to certain regions.\\
- It should only consider the bounding boxes that does appear in the image and not refer to any bounding boxes that actually do not exist. Note that all bounding boxes are numbered sequentially, so if we have 4 bounding boxes in the image, there are exactly the following bounding boxes: `R1`, `R2`, `R3`, and `R4`.\\
- It should list specific visual cues noticed within each highlighted region and how these observations lead to identifying the relevant subject (e.g., a man with a backpack).\\
- It should clearly connect each observation to the final answer. It should explain how, based on the evidence gathered from the image, the conclusion was reached (e.g., identifying the man holding a cell phone).\\
- If you realize you wrote something wrong in your thought process previously, you should identify and correct the mistakes you have made.\\

\\
\\
Write as if you are this person, in a similar style as above examples.
\\
\\
\#\# Image \texttt{\{im\_num\}} (\texttt{\{num\_bbox\}} bounding box(es))

**Question**: \texttt{\{question\}}\\
**Thought Process**: \\
**Answer**: \\
\\
\texttt{\{generation\_prompt\}}\\
\\
\texttt{\color{orange}\{false\_reasoning\}}\\
\hline
\end{tabularx}
\vspace{-1em}
\caption{Prompt for self correction. \texttt{\color{orange}\{false\_reasoning\}} is the injected incorrect sub-reasoning.}
\label{fig:prompt_self_correction}
\end{figure}

\begin{figure}[!htpb]
\centering
\fontsize{7}{9}\selectfont
\begin{tabularx}{1.0\columnwidth}{X}
\hline
**Evaluation Task Instructions:**\\

You will be given three pieces of information:\\
1. A **Question**\\
2. A **Model Output** (the answer and rationale generated by a language model)\\
3. A **Ground Truth Answer** (the correct answer)\\
\\
Your task is to find model's final conclusion of the question and determine if the final conclusion from **Model Output** is essentially equivalent to the **Ground Truth Answer** and correctly answers the **Question**. Consider the following guidelines when evaluating:\\
- If the final conclusion includes extra details, omissions, or slight wording differences but the overall meaning and essential information match the ground truth, consider it **Correct**.\\
- If the final conclusion provides incorrect information, adds unrelated details, or misses critical parts of the ground truth, consider it **Incorrect**.\\
\\
**Output Format (follow this exactly):**\\
\\
```\\
Reasoning: [Provide a brief explanation of why you judged the answer as Correct or Incorrect. Be explicit about which details match or differ.]\\
Correctness: [Write only one word: Either "Correct" or "Incorrect"]\\
```\\
\\
\texttt{\color{cyan}**In-Context Learning examples go here**}\\
\\
**Instance to Evaluate:**\\
*Question:* \texttt{\{question\}}\\
*Model Output:* \texttt{\{model\_output\}}\\
*Ground Truth:* \texttt{\{ground\_truth\}}\\
\hline
\end{tabularx}
\vspace{-1em}
\caption{Prompt for long-form answer judgment.}
\label{fig:prompt_full_answer}
\end{figure}

\begin{figure}[!htpb]
\centering
\fontsize{7}{9}\selectfont
\begin{tabularx}{1.0\columnwidth}{X}
\hline
You will be given three pieces of information:\\
1. A **Question**\\
2. A **Model Output** (the answer generated by the model)\\
3. A **Ground Truth Answer** (the correct answer)\\
\\
**Your task**: Determine if the **Model Output** correctly answers the **Question** by checking whether it aligns or essentially matches the **Ground Truth Answer**, considering meaning rather than exact wording.\\
\\
Follow these expanded guidelines carefully:\\
\\
- Mark as **Correct** if the meaning or essential details in the Model Output closely align with the Ground Truth Answer, even if different wording, synonyms, or minor descriptive variations are used.\\
\\
  - **Example**: "Beige," "tan," "cream," or "light brown" are considered close enough to be marked as **Correct**.\\
  - Similarly, minor differences like "couch" vs "sofa," or "silver-colored" vs "metallic grey" should still be considered **Correct** if they describe essentially the same thing.\\
- Mark as **Incorrect** only if the Model Output significantly differs from the Ground Truth Answer, changing the core meaning or providing substantially different or contradictory information.\\
\\
    - **Example**: If Ground Truth Answer is "wood," and Model Output is "metal," this would be marked **Incorrect** due to clear contradiction.\\
\\
**Output Format (follow this exactly):**\\
\\
```\\
Reasoning: [Provide a brief explanation of why you judged the answer as Correct or Incorrect. Be explicit about which details match or differ.]\\
Correctness: [Write only one word: Either "Correct" or "Incorrect"]\\
```\\
\\
\texttt{\color{cyan}**In-Context Learning examples go here**}\\
\\
**Instance to Evaluate:**\\
*Question:* \texttt{\{question\}}\\
*Model Output:* \texttt{\{model\_output\}}\\
*Ground Truth:* \texttt{\{ground\_truth\}}\\
\hline
\end{tabularx}
\vspace{-1em}
\caption{Prompt for short-form answer judgment.}
\label{fig:prompt_short_answer}
\end{figure}

\begin{figure*}[!htpb]
\centering
\fontsize{7}{9}\selectfont
\begin{tabularx}{1.0\textwidth}{X}
\hline
\# Task Description:\\
\\
You will be provided with three distinct pieces of information related to a reasoning task performed by another model:\\
\\
1. **Initial Reasoning**: An incomplete original response provided by a model containing incorrect reasoning.\\
2. **Correction**: A concise statement explicitly pointing out the specific mistake or incorrect reasoning in the Initial Reasoning and providing the correct information.\\
3. **Model's Continued Reasoning**: The model's new attempt at correcting the previously identified incorrect reasoning while complete the initial reasoning.\\
\\
Your job is to carefully evaluate whether the **Model's Continued Reasoning** successfully identified and corrected the incorrect reasoning based on the provided **Correction**.\\
\\
---\\
\\
\# Evaluation Steps:\\
\\
When evaluating, strictly follow these steps:\\
\\
\#\#\# Step 1: Identify the Incorrect Reasoning\\
\\
* Carefully read the **Initial Reasoning** and note exactly what mistake was made based on the provided **Correction**.\\
\\
\#\#\# Step 2: Verify the Provided Correction\\
\\
* Clearly understand the correction, noting specifically which detail from the **Initial Reasoning** was incorrect and what the accurate information is.\\
\\
\#\#\# Step 3: Evaluate the Model's Continued Reasoning\\
\\
* Read the **Model's Continued Reasoning** carefully to verify if the model explicitly identifies the original mistake.\\
* Ensure the model incorporates the corrected information provided by the **Correction**.\\
* Check whether the revised reasoning removes the previous mistake and accurately resolves it, not simply ignoring or sidestepping the original error.\\
\\
\#\#\# Step 4: Provide Clear Reasoning for Your Judgement\\
\\
* Explicitly state how the revised model output relates to both the initial incorrect reasoning and the provided correction.\\
* Clearly explain whether or not the original error was addressed and resolved.\\
\\
\#\#\# Step 5: Final Judgement\\
\\
* Clearly state your final judgement as either **"YES"** or **"NO"**:\\
\\
  * **"YES"** if the revised output explicitly corrects the initial error based on the provided correction.\\
  * **"NO"** if the revised output either does not identify the initial mistake explicitly, fails to correct it, or leaves the incorrect reasoning intact.\\
\\
---\\
\texttt{\color{cyan}**In-Context Learning examples go here**}\\
---\\
\# Final Notes for Your Judging:\\
\\
* Be meticulous and explicit in your reasoning.\\
* Your evaluation must precisely check whether the revised output aligns correctly with the provided correction.\\
* Ensure the final judgement ("Yes" or "No") directly and clearly corresponds to your detailed reasoning.\\
\\
---\\
\\
**Output Format (follow exactly):**\\
\\
```\\
Reasoning: [Provide a clear explanation of your judgment, explicitly mentioning the details or synonyms that align or differ.]\\
Correctness: [Write exactly one word: Either "Yes" or "No"]\\
```\\
\\
---\\
\\
**Instance to Evaluate:**\\
*Initial Reasoning:* \texttt{\{false\_reasoning\}}\\
*Correction:* \texttt{\{model\_output\}}\\
*Model's Continued Reasoning:* \texttt{\{ground\_truth\}}\\
\hline
\end{tabularx}
\vspace{-1em}
\caption{Prompt for self-correction judgment.}
\label{fig:prompt_judge_self_corr}
\end{figure*}

\end{document}